\documentclass[11pt,letterpaper]{article}
\usepackage[utf8]{inputenc}
\PassOptionsToPackage{hyphens}{url}\usepackage[hyperref]{emnlp2018}
\usepackage{times}
\usepackage{latexsym}
\usepackage{url}
\usepackage{soul}
\usepackage{subcaption}
\usepackage{textcomp}
\usepackage{graphicx}
\usepackage{geometry}
\usepackage{tikz}
\usepackage{amsmath}
\usepackage{amssymb}
\usepackage{booktabs}
\usepackage{siunitx}

\newcommand{\inline}[2]{%
    \begin{tikzpicture}[baseline=(word.base), txt/.style={shape=rectangle, inner sep=0pt}]
    \node[txt] (word) {#1};
    \node[above] at (word.north) {\tiny{#2}};
\end{tikzpicture}%
}
    
\aclfinalcopy 

\title{Team EP at TAC 2018: Automating data extraction in systematic reviews of environmental agents}
 
\author{Artur Nowak \and Paweł Kunstman \\
  Evidence Prime \\
  {\tt \{artur.nowak,pawel.kunstman\}@evidenceprime.com}
}
  
\date{}
\begin{document}
\maketitle
\begin{abstract}
We describe our entry for the Systematic Review Information Extraction track of the 2018 Text Analysis Conference.
Our solution is an end-to-end, deep learning, sequence tagging model based on the BI-LSTM-CRF architecture. However, we use interleaved, alternating LSTM layers with highway connections instead of the more traditional approach, where last hidden states of both directions are concatenated to create an input to the next layer.
We also make extensive use of pre-trained word embeddings, namely GloVe and ELMo.
Thanks to a number of regularization techniques, we were able to achieve relatively large capacity of the model (31.3M+ of trainable parameters) for the size of training set (100 documents, less than 200K tokens).
The system's official score was 60.9\% (micro-F1) and it ranked first for the Task 1. Additionally, after rectifying an obvious mistake in the submission format, the system scored 67.35\%.
\end{abstract}

\section{Introduction}
Systematic reviews play a fundamental role in health decision-making. They offer a comprehensive and unbiased synthesis of human knowledge on a given subject, produced through a standarized, transparent and scrupulous process. The result is crucial for creation of trustworthy health hazard assessments and clinical practice guidelines.

Systematic review process strives to achieve perfect recall. To accomplish that, the net is first cast wide -- thousands of papers are retrieved and are manually sifted. The included citations are then subject to data extraction, in which the information relevant to the given research question is selected. As advocated by organizations setting standards for systematic reviews, such as Cochrane, both steps should be performed independently by several researchers to rule out human errors or biases. This further increases the labor intensiveness of the process.

As a result, many reviews take up to two years to finish and may be already outdated at the moment of publication. The increase in the scientific output (more than 800,000 papers are indexed by MEDLINE every year), as well as questions requiring rapid responses (e.g. in cases of chemical spills), demand solutions for expediting the process without sacrificing quality. Recent advances in Natural Language Processing technologies may be the answer.

In 2018, the National Toxicology Program (NTP) and the Environmental Protection Agency (EPA) co-organized Systematic Review Information Extraction (SRIE) track as a part of Text Analysis Conference (TAC). The objective of the track was to evaluate automatic information extraction approaches that could aid in performing systematic reviews of environmental agents.

The track consisted of two tasks:

\begin{enumerate}
    \item Entity recognition of experimental design factors for the categories of exposure, animal group, dose group and endpoint.
    \item Relation extraction between experimental design factors from Task 1.
\end{enumerate}

This paper describes our approach for Task 1, the only one we participated in.

\section{Datasets and pre-processing}
\subsection{Characteristics of the datasets}

The organizers prepared two datasets: a training set consisting of 100 annotated "Material and methods" sections, extracted from PubMed Central articles, along with their identifiers; a test set of 100 such texts, for which annotation were not released to the participants until the submission deadline. To discourage manual annotation, the published test set contained further 344 texts that were not used in evaluation.

The training set contained nearly 7K sentences and 152K words (space-separated, our custom tokenization rules produced nearly 200K tokens). There were 15,253 mentions in the training set. Total of 24 entity classes were used, with the most frequent being: \textsc{Endpoint} (29\%), \textsc{TestArticle} (13\%) and \textsc{Species} (11\%). For six classes, there were less than 50 examples in the dataset: \textsc{TestArticleVerification} (6), \textsc{TimeAtLastDose} (23), \textsc{TestArticlePurity} (28), \textsc{CellLine} (39), \textsc{SampleSize} (45), \textsc{TimeAtFirstDose} (47).

Apart of the sheer number of classes, they also had complex semantics. For instance, the least frequent class (\textsc{TestArticleVerification}) was characterized in the annotation guidelines as: "Annotate the statement which indicates that the chemical was confirmed. This may refer to a third party
assessment where another company confirmed the chemical.". The decision boundary for even the most common classes is far from trivial. For example, \textsc{TestArticle} is defined as "the exposure (chemical or stressor) for which the experimental design is intended to evaluate [\ldots] Reagents used for endpoint analysis [\ldots] are not annotated as test articles.".

Furthermore, the mention endpoints often didn't agree with common tokenization rules. For instance, in cases like "15GD" (which is a shortcut for "15th gestational day"), "15" and "GD" were annotated as \textsc{TimeAtDose} and \textsc{TimeUnits}, respectively. In \textsc{GroupName} "Control-Sal", "Sal" was additionally annotated as \textsc{Vehicle} (saline).

This highlights another trait of the dataset: a large number of overlapping and discontiguous mentions. Among \textsc{Endpoint}s, 42\% of examples were discontiguous, i.e. they consisted of multiple spans that were not adjacent. Moreover, the discontiguous mentions often spanned multiple sentences. Furthermore, nearly 60\% of \textsc{Endpoint}s had at least one character in common with another \textsc{Endpoint} mention. Likewise, 16\% of \textsc{GroupName} mentions were discontiguous, similar number overlapped with another \textsc{GroupName} and more than twice the number overlapped with a \textsc{TestArticle}.

\subsection{Data pre-processing and augmentation} \label{sec:preprocessing}

\begin{figure*}[tb]
    \small
    \begin{subfigure}{\textwidth}
        \includegraphics[width=\textwidth]{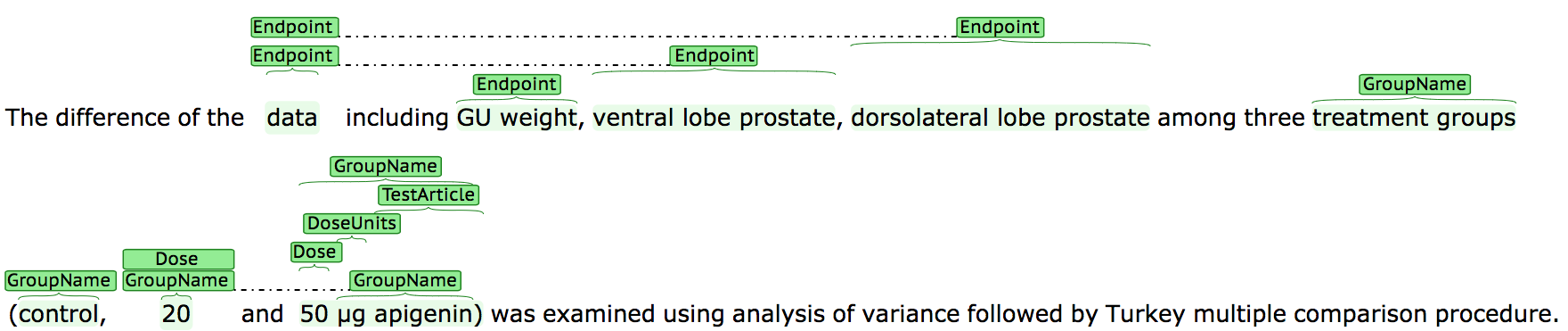}
        \caption{An example sentence, featuring both discontiguous and overlapping annotations}
    \end{subfigure}
    \begin{subfigure}{\textwidth}
    The difference between
    \inline{data}{B-Endpoint},
    including \inline{GU}{B-Endpoint}
    \inline{weight}{I-Endpoint},
    \inline{ventral}{B-Endpoint}
    \inline{lobe}{I-Endpoint}
    \inline{prostate}{I-Endpoint},
    \inline{dorsolateral}{B-Endpoint}
    \inline{lobe}{I-Endpoint}
    \inline{prostate}{I-Endpoint}
    between 3
    \inline{treatment}{B-GroupName}
    \inline{groups}{I-GroupName}
    (
    \inline{control}{B-GroupName}
    ,
    \inline{20}{B-GroupName}
    and
    \inline{50}{B-GroupName}
    \inline{\textmu g}{I-GroupName}
    \inline{apigenin}{I-GroupName}
    )
    was examined using the analysis of the variance followed by Turkey multiple comparison procedure.
    
    The difference between
    \inline{data}{B-Endpoint},
    including \inline{GU}{B-Endpoint}
    \inline{weight}{I-Endpoint},
    \inline{ventral}{B-Endpoint}
    \inline{lobe}{I-Endpoint}
    \inline{prostate}{I-Endpoint},
    \inline{dorsolateral}{B-Endpoint}
    \inline{lobe}{I-Endpoint}
    \inline{prostate}{I-Endpoint}
    between 3
    \inline{treatment}{B-GroupName}
    \inline{groups}{I-GroupName}
    (
    \inline{control}{B-GroupName}
    ,
    \inline{20}{B-Dose}
    and
    \inline{50}{B-Dose}
    \inline{\textmu g}{B-DoseUnits}
    \inline{apigenin}{B-TestArticle}
    )
    was examined using the analysis of the variance followed by Turkey multiple comparison procedure.
    \caption{Two sentences that jointly represent the input}
    \end{subfigure}
    \caption{An illustration of the data representation}
    \label{fig:encoding}
\end{figure*}

\begin{figure*}[tb]
    \small
    \begin{subfigure}[t]{.315\textwidth}
        Briefly, MWCNT trace metal contamination was 0.78\%, with sodium (0.41\%) and iron (0.32\%) being the major metal contaminants. \hl{Average} MWCNT surface area measured by nitrogen \hl{absorption-desorption technique} (Brunauer-Emmett-Teller method, BET) was 26 m2/g. MWCNT median length was 3.86 \textmu m and \hl{count mean diameter} was 49 ± 13.4 (\hl{mean} ± S.D.) nm, as determined by scanning electron microscopy of MWCNTs suspended in dispersion medium as described below [12].
        \caption{The original text}
    \end{subfigure}
    \hfill
    \begin{subfigure}[t]{.315\textwidth}
        In short, MWCNT contamination of trace metals was 0.78\%, sodium (0.41\%) and iron (0.32\%) being the main metal contaminants. The \hl{mean} surface area MWCNT measured by the nitrogen absorption-desorption technique (Brunauer-Emmett-Teller, BET) was 26 m2/g. MWCNT The median length was 3.86 \textmu m and the \hl{average counting diameter} was 49 ± 13.4 (\hl{Average} ± S.D.) NM, as determined by scanning electron microscopy of MWCNTs in suspension in a dispersion medium as described below [12].
        \caption{Translated to French and back to English}
    \end{subfigure}
    \hfill
    \begin{subfigure}[t]{.315\textwidth}
        Briefly, MWCNT trace metal contamination was 0.78\%, sodium (0.41\%) and iron (0.32\%) are the major metal contaminants. Average MWCNT surface area measured by \hl{adsorption-desorption} of nitrogen \hl{method} (brunauer-Emmett-teller method, Bet) was 26 m2/g. MWCNT average length was 3.86 \textmu m and \hl{calculate the average diameter} was 49 ± 13.4 (mean ± S. D.) nm, as determined using scanning electron microscopy MWCNTs suspended in the dispersion medium as described below [12].
        \caption{Translated to Russian and back to English}
    \end{subfigure}
    \caption{Example of the round-trip translation data augmentation technique at work. Overall, some degree of paraphrasing is achieved at the cost of incorrect replacements (absorption $\neq$ adsorption) in the most troublesome parts.}
    \label{fig:translations}
\end{figure*}

Texts were first broken into sentences using Punkt tokenizer from NLTK \cite{Punkt}. They were then split into tokens with spaCy library \cite{spacy2}, using a number of custom rules to ensure that token boundaries line up with the mention spans from the training set.

The spans were also stripped of trailing and leading white-space and punctuation (we observed that both were inconsistently used in the training annotations). In the cases, in which the offsets didn't match the tokens (e.g. due to inconsistent handling of new line characters in the training files or annotations that began mid-word), they were corrected manually.

Although, as previously mentioned, Task 1 involved a large number of overlapping and discontiguous entities, for this year's evaluation we focused on predicting 'linear' mentions. While many techniques exist for handling both overlapping and discontiguous mentions -- some of them were explored as part of TAC 2017 Adverse Drug Reaction Extraction from Drug Labels track \cite{ADRoverview} -- only some address mentions spanning multiple sentences. This problem is structurally similar to co-reference resolution, so perhaps methods used for this task can be explored in the future. Nevertheless, we feel that creating a robust model for the simplified problem is still a prerequisite for tackling the full challenge.

Furthermore, the official evaluation metric (micro-averaged F1) was calculated using partial matches with the gold annotations. The threshold was initially set to 40\%, with 50\% used in the final scoring. This effectively meant that it sufficed to detect just the longest segment for the most common case: multiple discontiguous \textsc{Endpoint}s that shared some 'prefix', but all had unique 'root' word, e.g. "genes [\ldots] functions", "genes [\ldots] processes".

Thus, we decided to transform the task into a sequence tagging problem using IOB2 (i.e. \textsc{B-} tag is used in the beginning of every mention) annotation scheme.
Discontiguous mentions were joined if the distance between consecutive spans was $\leq 5$ characters. The remaining ones were treated as separate mentions (with the same class).

If a sentence contained overlapping mentions, it was emitted for every 'level' of mentions with different classes. In other words, if there were $n$ classes associated with a token, the sentence was outputted $n$ times. First, mentions with the lowest number of spans and the greatest total length were chosen, then the ones that appeared second in such order etc. Figure~\ref{fig:encoding} illustrates the encoding.

Finally, we did a round-trip (or: back and forth) translations of the training text \cite{TranslationAugmentation}: through French (using Microsoft Translator API) and Russian (using Yandex API). The major difficulty in performing this task was preserving word alignments, so the mentions' character offsets could be restored. Hence, we replaced the mentions with encoded, untranslatable strings during the translation phase. This way, their position in the result could be easily determined. For total of 3 documents it wasn't possible to recover the offsets, so the original texts were used. This way, we were able to triple the number of training documents. Figure~\ref{fig:translations} shows this technique in action. Neither extra sentences for overlapping mentions, nor round-trip translations were used to evaluate out-of-fold predictions during cross-validation.

Each token was also supplemented with its relative position in document, rounded to two decimal places. For one of the runs, we downloaded paper titles and abstracts using PubMed API. We then identified all the abbreviations in the training text and the abstract using a rule-based algorithm \cite{Abbrs}. For all the tokens (that weren't stop words) we added a Boolean feature for whether they (or their expansions) appear in the paper title. We refer to this feature as \texttt{in\_title} in the description below.

\section{Model architecture and training} \label{sec:model}

\begin{figure}[htb]
    \includegraphics[width=.5\textwidth]{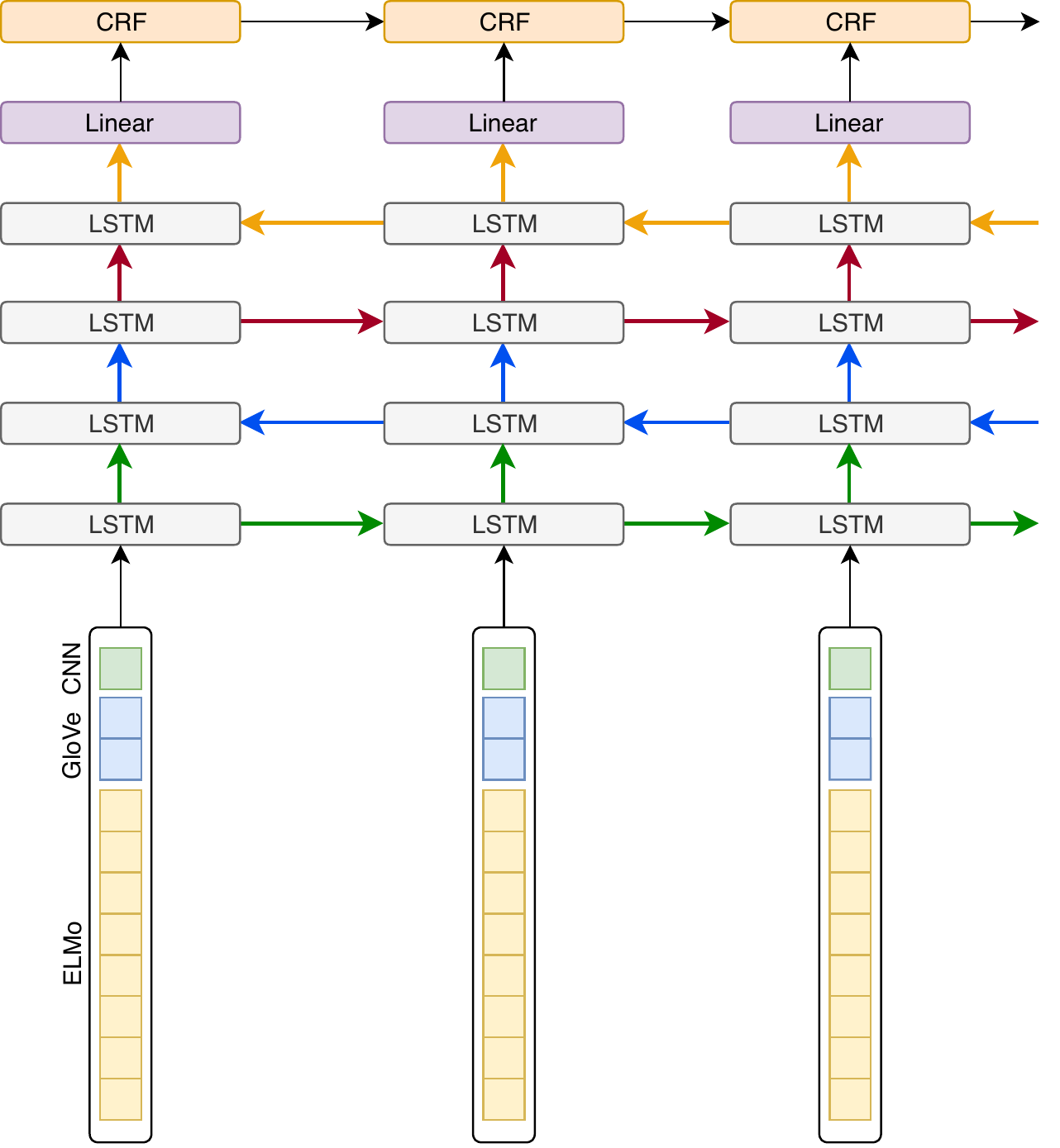}
    \caption{The model architecture. Coloured arrows share dropout masks.}
    \label{fig:architecture}
\end{figure}

Our solution is an end-to-end, deep learning, sequence tagging model based on the BI-LSTM-CRF \cite{HuangXY15} architecture. However, we use stacked, alternating LSTM layers \cite{AlternatingLSTM} with highway connections \cite{He2017DeepSR, Highway} instead of the more traditional approach, where last hidden states of both directions are concatenated to create an input to the next layer. The implementation was based on the AllenNLP library \cite{AllenNLP}. The model architecture is shown in Figure~\ref{fig:architecture}.

\textbf{Embeddings layer:} Each token is represented by 1452 dimensional vector, consisting of:

\begin{itemize}
    \item 300-dimensional GloVe \cite{pennington2014glove} embedding (cased, trained on 840B tokens from Common Crawl).
    \item 1024-dimensional ELMo \cite{ELMo} embedding that was originally trained on 5.5B tokens.
    \item 128-dimensional output from a character-level, one-layer CNN.
\end{itemize}

We fine-tuned ELMo vectors for two epochs on 48,141 PubMed abstracts (full texts, where available, total of nearly 365M tokens; training and test documents  were explicitly excluded). This resulted in drop in perplexity (measured on the Task 1 training set) from more than 300 to 27. This step is especially important, because of the custom tokenization rules that we mentioned in Section~\ref{sec:preprocessing} -- ELMo contains a character-level encoder.

Generally, ELMo exposes hidden state from all three layers of the model. The final representation is a weighted sum of these three vectors, where the weights are learned during training for the downstream task. However, in this work we set the L2 regularization parameter $\lambda = 1$ for the weights, which effectively leads to a simple average over the layers. Both GloVe and ELMo embeddings were frozen during training of the model.

The input to the CNN is 16-dimensional (learned) embedding of token characters. It is then passed to one convolution layer with kernel size of 3 (i.e. trigrams) and 128 filters. ReLU was used as the activation function. The concatenated embeddings were followed by a dropout layer (with $d_1 = 0.75$). Then, the remaining features were added -- in case of our best run, it was only the relative offset information.

\textbf{LSTM layers:} The token representations are then fed into four, alternating layers of LSTM with highway connections and hidden state of size 800. Highway connections \cite{Highway} are essentially an extension to the LSTM \cite{LSTM} that adds gated combination of the 'traditional' LSTM cell output and a linear transformation of its input:
\begin{align}
    h_t &= w_t \odot \tilde{h}_t + (1-w_t) \odot W_h\, [x_t] \label{eq:lstm_highway} \\
    \tilde{h}_t &= o_t \odot \tanh(c_t) \label{eq:lstm_old} \\
    c_t &= f_t \odot c_{t-1} + i_t \odot \tilde{c}_t \nonumber \\
    \tilde{c}_t &= \tanh(W_c\, [1; h_{t-1}; x_t]) \nonumber \\
    w_t &= \sigma(W_w\, [1; h_{t-1}; x_t]) \label{eq:highway_gate} \\
    o_t &= \sigma(W_o\, [1; h_{t-1}; x_t]) \nonumber \\
    f_t &= \sigma(W_f\, [1; h_{t-1}; x_t]) \nonumber \\
    i_t &= \sigma(W_i\, [1; h_{t-1}; x_t]) \nonumber
\end{align}
where $\odot$ is element-wise multiplication, $\sigma(\cdot)$ is element-wise sigmoid function and $[1; a; b]$ is a vector created by horizontally stacking vectors $[1]$ (for a bias term), $a$ and $b$. Here, $\tilde{h}_t$~(\ref{eq:lstm_old}) is the hidden state (time-step output) of the LSTM cell, as originally defined. Equation~\ref{eq:lstm_highway} is the new definition of the hidden state that uses the new gate defined in~\ref{eq:highway_gate}.

Moreover, $d_2 = 0.5$ dropout with a constant (i.e. the same for all time steps) mask is applied to $h_t$. This realizes variational inference based dropout introduced in \cite{VarDropout}. The only difference is that the same mask is used both for the 'output' (going to the next layer) and recurrent connections:
$$
  h'_t = z \odot h_t
$$

\textbf{CRF}: Traditional \cite{Dropout} $d_3 = 0.75$ dropout is applied to the last hidden state of the LSTM and the result is passed to a linear layer that projects the vector into a 49-dimensional (number of tags) space.

This vector is used as input to a linear-chain Conditional Random Field \cite{CRF}. The Viterbi algorithm is used for decoding, with constraints in place to penalize disallowed tag transitions (e.g. "\textsc{B-Endpoint} $\rightarrow$ \textsc{I-TestArticle}"). 

The model was trained to minimize negative log likelihood produced by the CRF using Adam optimizer \cite{Adam}, with starting learning rate of 0.001 and batch size of 32. 5-fold cross validation was used, with validation metric being micro-averaged F1 using exact span matching. The learning rate was halved if the validation metric didn't improve by at least 0.001 in 5 epochs. Early stopping was used with 10-epoch patience.

\section{Results}

The official evaluation script scored the submissions by creating a maximum match between gold annotations and predictions that intersect with them at least in 50\% (character-wise) and calculating micro-averaged F1 measure for such mapping. We submitted 3 runs: 
\begin{enumerate}
    \item Result of the model described in Section~\ref{sec:model} (without the \texttt{in\_title} feature), trained on the whole training set for 23 epochs.
    \item A majority vote of models (without the \texttt{in\_title} feature) trained during 5-fold CV. In case of ties, the first tag in lexicographic order was taken.
    The F1 score (using the official evaluation script) of the out-of-fold predictions was 69.90\% for 50\% similarity threshold.
    \item A majority vote of 10 models: five from the run 2 and another five, using \texttt{in\_title} feature, trained during 5-fold CV. The local CV score for the 5 \texttt{in\_title} models was 70.11\% for 50\% threshold.
\end{enumerate}

The submission files contained lists of mentions, along with the detected character offsets. There was some confusion how to count the line endings for the purpose of offset calculation. Our first submission treated all line endings (including Windows-style CRLF) as a single character. After contacting the organizers, this turned out to be incorrect. Our second and final submission counted all line endings as two characters, overlooking the fact that 10 out of 100 test files used single-character, Unix-style line endings.

We feel that the score obtained after rectifying this obvious mistake is more representative of the overall system performance. Therefore, we report both the official score (from our second submission) and the result of re-scoring our second submission after replacing these 10 files with the ones from our first submission. The results are presented in Tables~\ref{table:scores} and~\ref{table:best_run_full}.

\begin{table}[h]
\centering
\begin{tabular}{lcc}
\toprule
Run ID & Official score & Score with correction  \\
\midrule
\texttt{ep\_1} & $60.29$ & $66.76$ \\
\texttt{ep\_2} & $\mathbf{60.90}$ & $\mathbf{67.35}$ \\
\texttt{ep\_3} & $60.61$ & $67.07$ \\
\bottomrule
\end{tabular}
\caption{The scores of our three submitted runs for similarity threshold 50\%.}
\label{table:scores}
\end{table}

\begin{table}[h]
\centering
\small
\begin{tabular}{lS[table-format=5.0]S[table-format=2.2]S[table-format=2.2]}
 \toprule
 Mention class & {No. examples} & {F1 (5-CV)} & {F1 (Test)} \\
 \midrule
Total & 15265 & 69.90 & 67.35 \\ 
\midrule
Endpoint & 4411 & 66.89 & 61.47 \\
TestArticle & 1922 & 63.29 & 64.19\\
Species & 1624 & 95.33 & 95.95\\
GroupName & 963 & 67.08 & 62.40\\
EndpointUnitOfMeasure & 706 & 42.27 & 40.41\\
TimeEndpointAssessed & 672 & 57.27 & 55.51\\
Dose & 659 & 78.47 & 75.85\\
Sex & 612 & 96.27 & 98.36\\
TimeUnits & 608 & 68.03 & 61.26\\
DoseRoute & 572 & 69.24 & 69.80\\
DoseUnits & 493 & 77.50 & 72.33\\
Vehicle & 440 & 63.03 & 67.15\\
GroupSize & 387 & 77.79 & 75.74\\
Strain & 375 & 78.56 & 76.00\\
DoseDuration & 216 & 59.78 & 56.80\\
DoseDurationUnits & 204 & 57.83 & 56.60\\
TimeAtDose & 117 & 34.29 & 35.68\\
DoseFrequency & 96 & 41.56 & 59.78\\
TimeAtFirstDose & 47 & 3.92 & 0.00\\
SampleSize & 45 & 43.84 & 50.00\\
CellLine & 39 & 50.00 & 50.77\\
TestArticlePurity & 28 & 34.04 & 60.00\\
TimeAtLastDose & 23 & 0.00 & 0.00\\
TestArticleVerification & 6 & 0.00 & 0.00\\
\bottomrule
\end{tabular}
\caption{Detailed results of our best run (after correcting the submission format), along with numbers of mentions in the training set.}
\label{table:best_run_full}
\end{table}

\begin{table*}[htb]
\centering
\begin{tabular}{lS[table-format=2.2]S[table-format=1.2]S[table-format=2.2]S[table-format=1.2]S[table-format=2.2]S[table-format=1.2]}
\toprule
ID & {5-fold CV} & {$\Delta$} & {Single model} & {$\Delta$} & {Ensemble} & {$\Delta$} \\
\midrule
LSTM-800 & 70.56 & +0.66 & 67.54 & +0.78 & 67.65 & +0.30 \\
LSTM-400 & 70.50 & +0.60 & $\mathbf{67.59}$ & +0.83 & $\mathbf{68.00}$ & +0.65 \\
IN-TITLE & 70.11 & +0.21 & {N/A} & {N/A} & 67.52 & +0.17 \\
\textbf{SUBMISSION} & 69.90 & {--} & 66.76 & {--} & 67.35 & {--} \\
NO-HIGHWAY & 69.72 & -0.18 & 66.42 & -0.34 & 66.64 & -0.71 \\
NO-OVERLAPS & 69.46 & -0.44 & 65.07 & -1.69 & 66.47 & -0.88 \\
LSTM-400-DROPOUT & 69.45 & -0.45 & 65.53 & -1.23 & 67.28 & -0.07 \\
NO-TRANSLATIONS & 69.42 & -0.48 & 65.92 & -0.84 & 67.23 & -0.12 \\
NO-ELMO-FINETUNING & 67.71 & -2.19 & 65.16 & -1.60 & 65.42 & -1.93 \\
\bottomrule
\end{tabular}
\caption{The estimation of impact of various design choices on the final result.  The entries are sorted by the out-of-fold scores from CV. The \textbf{SUBMISSION} here uses score from \texttt{ep\_1} run for the single model and \texttt{ep\_2} for the ensemble performance.}
\label{table:ablations}
\end{table*}

\section{Discussion}

To better estimate the impact of the described techniques on our final result, we performed a series of ablation studies. We re-trained our best model (\texttt{ep\_2}), removing one of its feature at time:
\begin{enumerate}
    \item \textbf{LSTM-800}: Stacked BI-LSTM with two layers and hidden size of 800 (instead of four alternating LSTM layers).
    \item \textbf{LSTM-400}: Stacked BI-LSTM with two layers of size 400.
    \item \textbf{IN-TITLE}: Majority vote of 5 \texttt{in\_title} models -- in other words, \texttt{ep\_3} submission without ensembling with \texttt{ep\_2} models.
    \item \textbf{NO-HIGHWAY}: Traditional LSTM cell definition, without the highway connection.
    \item \textbf{NO-OVERLAPS}: Without extra sentences generated for overlaps (as in Figure~\ref{fig:encoding}).
    \item \textbf{LSTM-400-DROPOUT}: Stacked BI-LSTM with two layers of size 400 and dropout only between LSTM layers, as proposed in \cite{TraditionalRnnDropout}.
    \item \textbf{NO-TRANSLATIONS}: Without the round-trip translation augmentation (see Figure~\ref{fig:translations}).
    \item \textbf{NO-ELMO-FINETUNING}: ELMo vectors as published by \cite{ELMo}, without fine-tuning on the PubMed data.
\end{enumerate}

To enable fair comparison of the ablated models with the submitted ones, we trained them both for 23 epochs (as was the case with the \texttt{ep\_1} submission) and during 5-fold cross-validation, using the same learning rate schedule and early stopping strategy, ensembling the results the same way as was done for the \texttt{ep\_2}. The results are presented in Table~\ref{table:ablations}.

Perhaps the most striking thing about the ablation results is that the 'traditional' LSTM layout outsperformed the 'alternating' one we chose for our submission. Our decision at that time was based on a score calculated on 20\% validation set, due to time constraints. Cross-validation results clearly show the winner here, although their translation into performance on the test set is inconsistent. Apart of the flipped results of the LSTM-800 and the LSTM-400, small differences in CV score are sometimes associated with large discrepancies in test set performance. This is mostly due to small size of the data set (low precision of the estimate), stochastic nature of the training process and hyperparameters (such as number of epochs for single models) aligning better with some of the models.

The results of ensembling are also varied. More testing is required to pinpoint the actual impact on the final score, by re-training the models several times with different random seeds and averaging the results. Some ablated models that perform poorly in the single-model scenario (e.g. NO-OVERLAPS, LSTM-400-DROPOUT) are able to regain a lot of accuracy when ensembled. Also, our data augmentation technique (NO-TRANSLATIONS) seem to have far smaller impact on the final score then we expected. Finally, the fact that our third run (\texttt{ep\_3}) fared worse than the second one is explained by the fact that we needlessly included the models from \texttt{ep\_2} in the ensemble (IN-TITLE). This makes us optimistic about future work on including title information in the model.

We presented our solution for automating data extraction in systematic reviews of environmental agents. Although we made considerable simplifications to the original problem setting, our error analysis shows that the system already delivers substantial value for potential users.

We plan to further explore the problem of structured predictions, taking into the consideration discontiguous and overlapping mentions. We also would like to apply the experience we gained from working on this problem to the Task 2 of the Track.

Our current best model completely ignores the document context. This information is intuitively the single most important feature for many classes, as the whole articles will generally describe the same \textsc{Specie}s, \textsc{TestArticle}s etc.

However, our extensive exploration of adding the document context by means of attention layers, carrying over hidden state, memory cells or even hand-crafted features led to unsatisfactory results. We think that the major cause of this situation is overfitting, due to a small number of documents. We also plan to work on alleviating this issue in our further work.

\bibliography{ep_tac_srie2018}
\bibliographystyle{acl_natbib_nourl}
\end{document}